\documentclass[runningheads]{llncs}
\usepackage{graphicx}
\usepackage{tikz}
\usepackage{comment}
\usepackage{amsmath,amssymb} % define this before the line numbering.
\usepackage{color}

\usepackage[accsupp]{axessibility}  % Improves PDF readability for those with disabilities.

\usepackage[pagebackref,breaklinks,colorlinks]{hyperref}
\usepackage[capitalize]{cleveref}
\crefname{section}{Sec.}{Secs.}
\Crefname{section}{Section}{Sections}
\Crefname{table}{Table}{Tables}
\crefname{table}{Tab.}{Tabs.}
\Crefname{algocf}{Algorithm}{Algorithms}

\usepackage[linesnumbered,ruled,vlined]{algorithm2e}
\usepackage{threeparttable}
\usepackage{booktabs}
\usepackage{multirow}
\usepackage{caption}
\usepackage{subcaption}
\usepackage{appendix}
\usepackage{hyperref}

\usepackage{xcolor}

\usepackage{amsmath}
\DeclareMathOperator*{\argmax}{arg\,max}
\DeclareMathOperator*{\argmin}{arg\,min}

\begin{document}
% \linenumbers
\pagestyle{headings}
\mainmatter
\def\ECCVSubNumber{6978}  % Insert your submission number here

\title{LA3: Efficient Label-Aware AutoAugment} % Replace with your title

% \begin{comment}
\titlerunning{LA3}
\authorrunning{M. Zhao et al.} 
\author{Mingjun Zhao\inst{1}, Shan Lu\inst{1}, Zixuan Wang\inst{2}, Xiaoli Wang\inst{2} and Di Niu\inst{1}}
\institute{University of Alberta \and Platform and Content Group, Tencent\\
\email{\{zhao2,slu1,dniu\}@ualberta.ca}, 
\email{\{zackiewang,evexlwang\}@tencent.com}}

% \end{comment}
%******************

\maketitle

\begin{abstract}
Automated augmentation is an emerging and effective technique to search for data augmentation policies to improve generalizability of deep neural network training. Most existing work focuses on constructing a unified policy applicable to all data samples in a given dataset, without considering sample or class variations.
In this paper, we propose a novel two-stage data augmentation algorithm, named \textit{Label-Aware AutoAugment (LA3)}, which takes advantage of the label information, and learns augmentation policies separately for samples of different labels. 
\textit{LA3} consists of two learning stages, where in the first stage, individual augmentation methods are evaluated and ranked for each label via Bayesian Optimization aided by a neural predictor, which allows us to identify effective augmentation techniques for each label under a low search cost. 
And in the second stage, a composite augmentation policy is constructed out of a selection of effective as well as complementary augmentations, which produces significant performance boost and can be easily deployed in typical model training.
Extensive experiments demonstrate that \textit{LA3} achieves excellent performance matching or surpassing existing methods on CIFAR-10 and CIFAR-100, and achieves a new state-of-the-art ImageNet accuracy of $79.97\%$ on ResNet-50 among auto-augmentation methods, while maintaining a low computational cost.
\end{abstract}

\section{Introduction}

\begin{figure}[t]
\centering
\includegraphics[width=0.65\columnwidth]{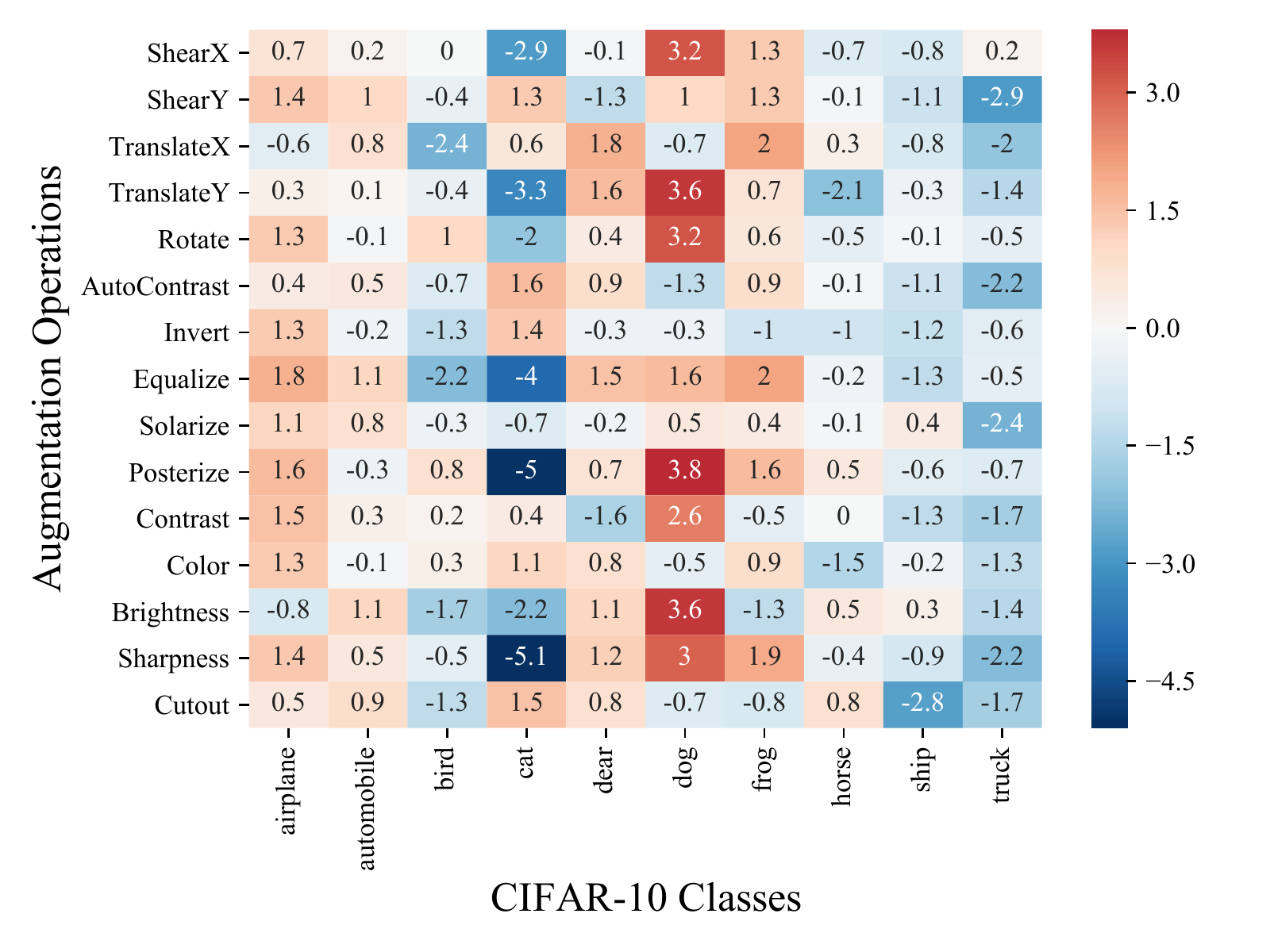}
\caption{The effects of different augmentation operations on each class in CIFAR-10, demonstrated by the test accuracy change in each class after each single augmentation is applied to training WRN-40-2.
}
\label{fig:Demo}
\end{figure}

Data augmentation has proven to be an effective regularization technique that can improve the generalization of deep neural networks by adding modified copies of existing samples to increase the volume and diversity of data used to train these networks. 
Traditional ways of applying data augmentation in computer vision include using single augmentation techniques, such as rotation, flipping and cutout \cite{devries2017improved}, adopting randomly selected augmentations \cite{cubuk2020randaugment}, and employing a manually crafted augmentation policy consisting of a combination of transformations. 
However, these methods either do not reach the full potential of data augmentation, or require human expertise in policy design for specific tasks.

Recently, automated learning of augmentation policies has become popular to surpass the limitation of manual design, achieving remarkable advances in both the performance and generalization ability on image classification tasks.
Different search algorithms such as reinforcement learning \cite{cubuk2019autoaugment}, population-based training \cite{ho2019population}, and Bayesian Optimization \cite{lim2019fast} have been investigated to search effective augmentation policies from data to be used to train  target networks.
Dynamic augmentation strategies, e.g., PBA \cite{ho2019population}, AdvAA \cite{zhang2019adversarial}, are also proposed to learn non-stationary policies that vary during model training.

However, most existing methods focus on learning a single policy that is applied to all samples in the dataset equally, without considering variations between samples, classes or labels, which may lead to sub-optimal solutions.
\Cref{fig:Demo} demonstrates the effects of different augmentation operations on different classes of samples in CIFAR-10, from which we can see that the effectiveness of augmentations is different on each class.
For example, when the operation ``Posterize'' is applied in training, the test accuracy of ``dog'' class increases by $3.8\%$, whereas the test accuracy of ``cat'' drops significantly by $5\%$. It is possible that a certain augmentation used in training has completely different impacts on different labels.
This observation implies the limitation of label or sample-invariant dataset-level augmentation policies.
MetaAugment \cite{zhou2021metaaugment} proposes to learn a sample-aware augmentation policy by solving a sample re-weighting problem. It uses an augmentation policy network to take an augmentation operation and the corresponding augmented image as inputs, and outputs a weight to adjust the augmented image loss computed by the task network.  
Despite the benefit of a fine-grained sample-dependent policy, MetaAugment is time-consuming and couples policy network learning with target model training, which may not be convenient in some production scenarios that require functional decomposition.

% Furthermore, in some popular methods such as AutoAugment \cite{cubuk2019autoaugment} and FastAA \cite{lim2019fast}, the final policy used to train target models is constructed by simply putting together the best performing augmentations, ignoring their complementary effects and most likely leading to the application of highly redundant operations.
% Other methods aim to learn non-stationary or dynamic policies, e.g., PBA \cite{ho2019population}, AdvAA \cite{zhang2019adversarial} and MetaAug \cite{zhou2021metaaugment}, that vary augmentation operations and policies during model training.
% \blue{However, they usually incur a high computational cost and require non-trivial modifications to the model training code, which is not always possible in production deep learning, where the functional decoupling of training recipes and augmentation policies is desirable for ease of deployment.}

In this paper, we propose an efficient data augmentation strategy named \textit{Label-Aware AutoAugment (LA3)}, which produces label-aware augmentation policies to overcome the limitation of sample-invariant augmentation while still being computationally efficient as compared to sample-aware or dynamic augmentation strategies. 
\textit{LA3} achieves competitive performance matching or outperforming a wide range of existing static and dynamic auto-augment methods, and attains the highest ImageNet accuracy on ResNet-50 among all existing augmentation methods including dynamic ones.
In the meantime, \textit{LA3} is also a simple scheme which separates augmentation policy search from target network model training, and produces stationary augmentation policies that can easily be applied to enhance deep learning with minimum perturbation to the original target model training routine.

\textit{LA3} adopts a two-staged design, which first explores a search space of combinations of operations and evaluates the effectiveness of promising augmentation operations for each class, while in the second stage, forms a composite policy to be used in target model training. 

In the first stage of \textit{LA3}, a neural predictor is designed to estimate the effectiveness of operation combinations on each class and is trained online through density matching as the exploration process iterates. We use Bayesian Optimization with a predictor-based sampling strategy to guide search into meaningful regions, which greatly improves the efficiency and reduces search cost. 
 
In the second stage, rather than only selecting top augmentation operations, we introduce a policy construction method based on the minimum-redundancy maximum-reward (mRMR) principle \cite{peng2005feature}
to enhance the performance of the composite augmentation policy when applied to the target model.
This is in contrast to most prior methods \cite{cubuk2019autoaugment}, \cite{lim2019fast}, which simply put together best performing augmentations in evaluation, ignoring their complementary effects.

Extensive experiments show that using the same set of augmentation operations, the proposed \textit{LA3} achieves excellent performance outperforming other low-cost static auto-augmentation strategies, including FastAA and DADA, on CIFAR-10 and CIFAR-100, in terms of the accuracy.
%and achieves a new state-of-the-art ImageNet accuracy of $79.97\%$ on ResNet-50 without extra data.
On ImageNet, \textit{LA3}, using stationary policies, achieves a new state-of-the-art top-1 accuracy of $79.97\%$ on ResNet-50, which outperforms prior auto-augmentation methods including dynamic strategies such as AdvAA and MetaAug, while being $2\times$ and $3\times$ more computationally efficient, respectively.

\section{Related Work}

%In this section, we give a brief review of the related work on automated data augmentations.

Data augmentation is a popular technique to alleviate overfitting and improve the generalization of neural network models by enlarging the volume and diversity of training data. 
Various data augmentation methods have been designed, such as Cutout \cite{devries2017improved}, Mixup \cite{zhang2018mixup}, CutMix \cite{yun2019cutmix}, etc. 
Recently, automated augmentation policy search has become popular, replacing human-crafted policies by learning policies directly from data.
AutoAugment \cite{cubuk2019autoaugment} adopts a reinforcement learning framework that alternatively evaluates a child model and trains an RNN controller to sample child models to find effective augmentation policies.
Although AutoAugment significantly improves the performance, its search process can take thousands of GPU hours which greatly limits its usability.

Multiple strategies are proposed to lower the search cost.
Fast AutoAugment \cite{lim2019fast} proposes a density matching scheme to avoid training and evaluating child models, and uses Bayesian Optimization as the search algorithm. 
Weight-sharing AutoAugment \cite{tian2020improving} adopts weight-sharing settings and harvests rewards by fine-tuning child models on a shared pre-trained target network.
Faster AutoAugment \cite{hataya2020faster} further reduces the search time by making the search of policies end-to-end differentiable through gradient approximations and targeting to reduce the distance between the original and augmented image distributions.
Similarly, DADA \cite{li2020dada} relaxes the discrete policy selection to a differentiable optimization problem via Gumbel-Softmax \cite{jang2016categorical} and introduces an unbiased gradient estimator.

Instead of producing stationary augmentation policies that are consistent during the target network training, PBA \cite{ho2019population} learns a non-stationary augmentation schedule, inspired by population based training \cite{jaderberg2017population}, by modeling the augmentation policy search task as a process of hyperparameter schedule learning.
AdvAA \cite{zhang2019adversarial} adopts an adversarial framework that jointly optimizes target network training and augmentation search to find harder augmentation policies that produce the maximum training loss.
However, AdvAA must rely on the batch augment trick, where each training batch is enlarged by multiple times with augmented copies, which significantly increases its computational cost.
In general, one concern of these dynamic strategies is that  they intervene the standard model training procedure, causing extra deployment overhead and may not be applicable in many production environments.

While most previous studies focus on learning augmentation policies for the entire dataset,
MetaAugment \cite{zhou2021metaaugment} proposes to learn sample-aware augmentation policies during model training by formulating the policy search as a sample re-weighting problem, and constructing a policy network to learn the weights of specific augmented images by minimizing the validation loss via meta learning.
Despite its benefits, MetaAugment is computationally expensive, requiring three forward and backward passes of the target network in each iteration.
LB-Aug \cite{wang2021fine} is a concurrent work that also searches policies dependent on labels, but focuses on a different task under multi-label scenarios, where each sample has multiple labels rather than a single classification label.
LB-Aug uses an actor-critic reinforcement learning framework and policy gradient approach for policy learning.
Despite the benefits from label-based policies, LB-Aug has potential stability issues due to the use of reinforcement learning, which is generally harder and computational costly to train. In fact, the search cost of LB-Aug is not reported.
In contrast, \textit{LA3} targets the classical single-label image classification tasks, e.g., on CIFAR-10/100 and ImageNet benchmarks, on which most other auto-augmentation methods are evaluated.
It adopts Bayesian Optimization coupled with a neural predictor to sample and search for label-dependent augmentation policies efficiently. In addition, a policy construction stage is proposed to further form a more effective composite policy for target network training.

\section{Methodology}

In this section, we first review the task of conventional augmentation search and introduce the formulation of the proposed label-aware augmentation search task.
Then we describe the two-stage design of \textit{LA3}, and present the algorithm in detail.

\subsection{Conventional Augmentation Search}

Given an image recognition task with a training dataset $D^{tr}=\{(x_i, y_i\}_{i=1}^{|D^{tr}|}$, with $x_i$ and $y_i$ representing the image and label respectively, augmented samples $\mathcal{T}(x_i)$ are derived by applying augmentation policy $\mathcal{T}$ to sample $x_i$.
Usually, the policy $\mathcal{T}$ is composed of multiple sub-policies $\tau$, and each sub-policy is made up by $K$ augmentation operations $O$, optionally with their corresponding probabilities and magnitudes, which are adopted in the original design of AutoAugment \cite{cubuk2019autoaugment}, but not included in some of the recent methods such as Weight-sharing AutoAugment \cite{tian2020improving} and MetaAugment \cite{zhou2021metaaugment}.

Conventional augmentation search methods focus on the task whose goal is to construct the optimal policy $\mathcal{T}^*$ from given augmentations so that the performance $\mathcal{R}$ of the task network $\theta_\mathcal{T}$ on the validation dataset $D^{\textit{val}}$ is maximized:
\begin{equation}
	\label{eq:dataset_goal}
	\begin{aligned}
	\mathcal{T}^* &= \argmax_\mathcal{T} \mathcal{R}(\theta_\mathcal{T}|D^{val}), \\
	\text{where}\quad\theta_\mathcal{T} &= \argmin_{\theta_\mathcal{T}} \frac{1}{|D^{tr}|} \sum_{i=1}^{|D^{tr}|} \mathcal{L}_\theta(\mathcal{T}(x_i), y_i),\\
	\end{aligned}
\end{equation}
and $\mathcal{L}_\theta$ is the loss function of target network $\theta$.

\begin{figure*}[t]
\centering
\includegraphics[scale=0.72]{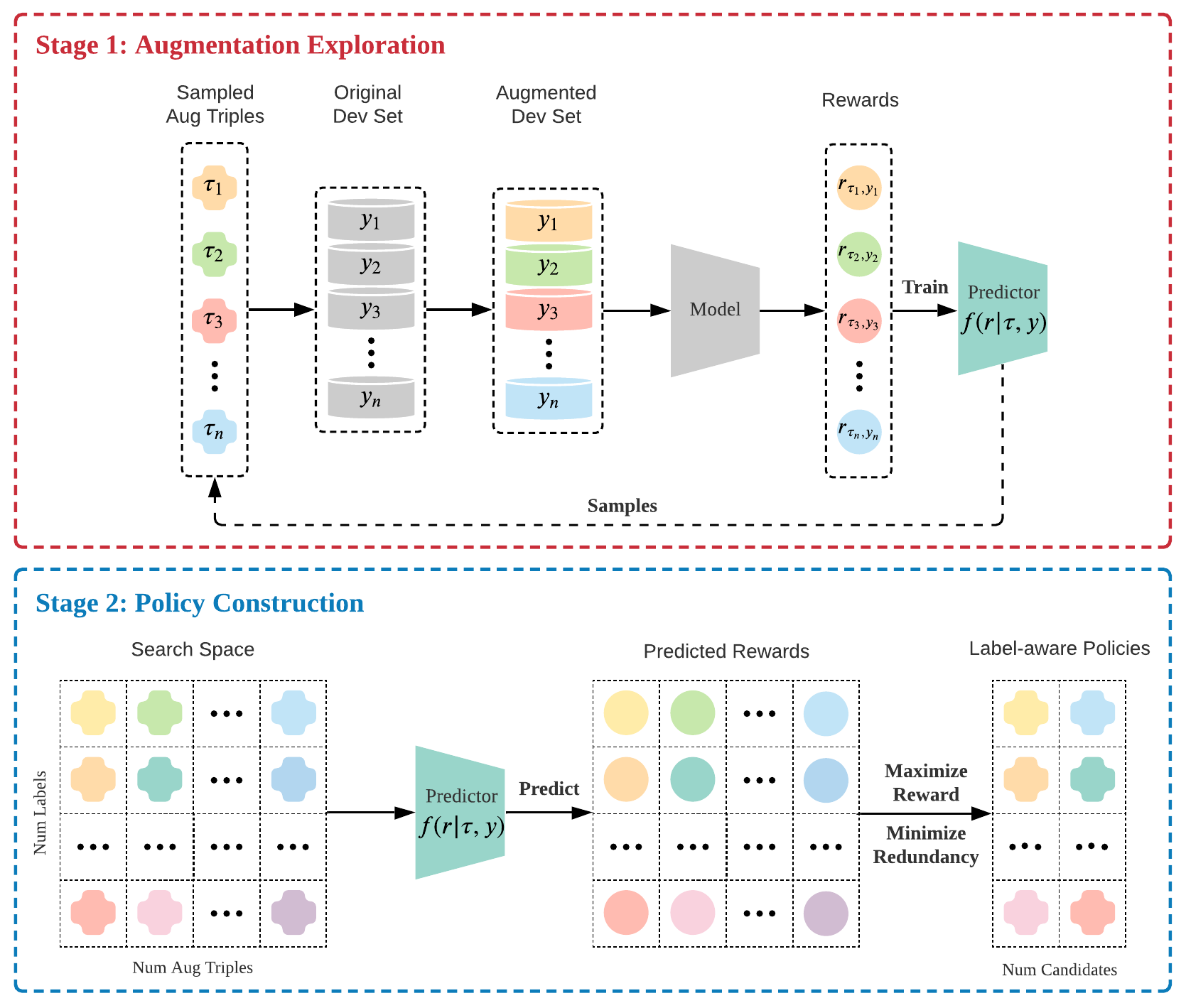}
\caption{An overview of the proposed \textit{LA3} method. It contains two stages, where in the first stage, augmentation triples are individually evaluated for each label via Bayesian Optimization with the help of an label-aware neural predictor. In the second stage, the best combination of complementary augmentation triples is selected based on the minimum-redundancy maximum-reward principle.
}
\label{fig:MethodOverview}
\end{figure*}

\subsection{Label-Aware Augmentation Search}
Though learning a dataset-level policy achieves considerable improvements, it is unlikely the optimal solution due to the lack of consideration of sample variations and utilization of label information.

In this paper, we aim to learn a label-aware data augmentation policy $\mathcal{T}^*=\{\mathcal{T}^*_{y_0}, \cdots, \mathcal{T}^*_{y_n}\}$, where for samples of each label $y_j$, an individual policy $\mathcal{T}_{y_j}$ is learned by maximizing the label-specific performance $\mathcal{R}_{y_j}$ of label $y_j$:
\begin{equation}
	\label{eq:label_goal}
	\begin{aligned}
	\mathcal{T}^*_{y_j} &= \argmax_{\mathcal{T}_{y_j}} \mathcal{R}_{y_j}(\theta_{\mathcal{T}}|D^{val}), \\
	\text{where}\quad\theta_{\mathcal{T}} &= \argmin_{\theta_{\mathcal{T}}} \frac{1}{|D^{tr}|} \sum_{i=1}^{|D^{tr}|} \mathcal{L}_\theta(\mathcal{T}_{y_i}(x_i), y_i).\\
	\end{aligned}
\end{equation}

Similar to conventional augmentation, in our label-aware setting, we define that each policy for a label is composed of multiple augmentation triples, each consisting of three augmentation operations. 
%We include in total $16$ image operations in our search space, including Identity, ShearX/Y, TranslateX/Y, Rotate, AutoContrast, Invert, Equalize, Solarize, Posterize, Contrast, Color, Brightness, Sharpness, and Cutout \cite{devries2017improved}. 
The magnitude of each augmentation operation is chosen randomly from ranges defined in AutoAugment \cite{cubuk2019autoaugment}, and is excluded from the search space in order to introduce randomness and diversity into the policy, and allocate more computational resources to assessing the fitness of operations to different classes of samples.
% control the size of the search space from being excessively large due to the label-aware setting.
%We do not search for magnitudes so that more computational resources can be allocated to assessing the fitness of operations to different classes of samples.

In this paper, we propose a label-aware augmentation policy search algorithm called \textit{LA3}, composed of two stages as presented in  \Cref{fig:MethodOverview}.
The first augmentation exploration stage aims to search for effective augmentation triples with density matching, and train a neural predictor to provide evaluations on all seen and unseen augmentation triples in the search space.
And the goal of the second policy construction stage is to build a composite policy for each label based on the evaluation results from stage 1 by selecting a subset of complementary augmentation triples based on the minimum-redundancy maximum-reward principle.
% The details of the algorithm is introduced in the following section.

\subsection{Stage 1: Augmentation Exploration}
\textbf{Density Matching} is an efficient mechanism originally proposed by Fast AutoAugment \cite{lim2019fast} to simplify the search process for effective augmentations, since the problem defined by \Cref{eq:dataset_goal} and \Cref{eq:label_goal} is a bi-level optimization problem, and is extremely hard to solve directly.
It calculates the reward of each augmentation triple without the need of repeatedly training the target network.
Specifically, given a model $\theta$ pre-trained on the training set $D^{tr}$ and a validation set $D^{val}$, the performance of a certain augmentation triple $\tau$ can be evaluated by approximately measuring the distance between the density of $D^{tr}$ and density of augmented validation set $\tau(D^{val})$ with the model performance $\mathcal{R}(\theta|\tau(D^{val}))$.
And the reward $r$ is measured by the performance difference caused by applying the augmentation triple $\tau$:
\begin{equation}
	\label{eq:density_dataset}
	r_\tau = \mathcal{R}(\theta|\tau(D^{val})) - \mathcal{R}(\theta|D^{val}).
\end{equation}
Similarly, in our label-aware setting, the reward $r$ for a certain augmentation triple $\tau_y$ at label $y$ is given by
\begin{equation}
	\label{eq:density_label}
	r_{\tau,y} = \mathcal{R}_y(\theta|\tau_y(D^{val})) - \mathcal{R}_y(\theta|D^{val}).
\end{equation}

\textbf{Bayesian Optimization with a Neural Predictor} is a widely adopted framework in many applications such as neural architecture search \cite{white2021bananas,wen2020neural} to find the optimal solution within a search space.
In standard BO setting, over a sequence of iterations, the results from previous iterations are used to model a posterior distribution to guide the candidate selection of next iteration.
And a neural predictor is a neural network that is repeatedly trained on the history evaluated candidates, and provides evaluations on unseen candidates, which increases the utilization efficiency of history evaluations and notably accelerates the search process.

In our \textit{LA3} algorithm, we incorporate a label-aware neural predictor $f(r|\tau,y)$ which takes in an augmentation triple $\tau$ and the label $y$ it is evaluated on, and predicts the reward $r$.
In each iteration, the sampled augmentation triples for different labels are evaluated according to \Cref{eq:density_label}, and together with the previous evaluated augmentation triples, are passed to train a new predictor.

\SetKwInput{KwInput}{Input}
\SetKwInput{KwOutput}{Output}
\SetArgSty{textnormal}

\begin{algorithm}[tb]
\caption{Stage 1: Augmentation Exploration}
\label{alg:stage1}
\DontPrintSemicolon
\KwInput{Pre-trained target network $\theta$, warm up iterations $T_0$, total iterations $T$}
\KwOutput{Well-trained predictor $f^T(r|\tau, y)$}

\tcc{warm-up phase}
\For{$t=0,\cdots,T_0$} 
{
	randomly generate augmentation triples $\{\tau^t_{y_0}, \cdots, \tau^t_{y_n}\}$ for all labels $\{y_0,\cdots,y_n\}$ \\
	obtain rewards $\{r_{\tau, y_0}^t,\cdots,r_{\tau, y_n}^t\}$ by  \Cref{eq:density_label} \\
}
	
\tcc{search phase}
\For{$t=T_0,\cdots,T$}
{
	train $f^t(r|\tau, y)$ with data collected from previous $t$ iterations $\{(\tau, y, r_{\tau, y})\}^t$\\
	\For{$y_i=y_0,\cdots,y_n$}
	{
		generate $100$ candidate augmentation triples by exploration and exploitation\\
		obtain predicted rewards $\tilde{r}_{\tau, y_i}=f^t(\tau, y_i)$ for $100$ candidates \\
		$\tau^t_{y_i}=\argmax_\tau(\tilde{r}_{\tau, y_i})$\\
	}

	obtain real rewards $\{r_{\tau,{y_0}}^t,\cdots,r_{\tau,{y_n}}^t\}$ for $\{\tau^t_{y_0}, \cdots, \tau^t_{y_n}\}$ by \Cref{eq:density_label}\\
}

train predictor $f^T(r|\tau, y)$ with all collected data $\{(\tau, y, r_{\tau, y})\}^T$\\
\end{algorithm}

Next, we select $100$ candidate augmentation triples at the balance of exploration and exploitation, based on the following selection procedure: 
1) Generate $10$ new candidates by randomly mutating $1$ or $2$ operations in the chosen augmentation triples of the previous iteration;
2) Randomly sample $50$ candidates from all unexplored augmentation triples;
3) Sample $40$ candidates from the explored augmentation triples according to their real reward values.
Then, for each label $y$, we choose the augmentation triple $\tau$ with the highest predicted reward $\tilde{r}_{\tau, y}$ for evaluation.

\textbf{Overall workflow} of the first stage is summarized in \Cref{alg:stage1}.
To begin with, a warm-up phase of $T_0$ iterations is incorporated to randomly explore the search space, and retrieve the initial training data for learning a label-aware neural predictor $f(r|\tau, y)$.
Then, for the following $T-T_0$ iterations, the search phase is adopted.
In each iteration, we first train a neural predictor from scratch with data collected from previous iterations.
Then, for each label, we apply the fore-mentioned selection procedure to select a set of candidate augmentation triples, and use the trained predictor to choose the augmentation triple for evaluation.
After enough training data is collected, a well-trained label-aware neural predictor can be derived to provide accurate evaluations on all augmentation triples for different labels.

\begin{algorithm}[tb]
\caption{Stage 2: Policy Construction}
\label{alg:stage2}
\DontPrintSemicolon
\KwInput{Well-trained predictor $f^T(r|\tau, y)$, search space $A$, number of candidates $N_\textit{cand}$}
\KwOutput{Label-aware policy $\mathcal{T^*}$}

\For{$y_i = y_0,\cdots,y_n$}
{
	\For{$\tau \in A$}
	{
		predict the reward $\tilde{r}_{\tau,y_i}=f^T(\tau, y_i)$\\
	}
	
	initialize label-specific policy $\mathcal{T}_{y_i}\leftarrow\emptyset$ \\
	\For{$k=0,\cdots,N_\textit{cand}$}
	{
		\For{$\tau \in (A\setminus\mathcal{T}_{y_i})$}
		{
			calculate $v(\tau,y_i)$ using \Cref{eq:score_function}\\
		}
		find augmentation triple with highest score $\tau^k=\argmax_{\tau}(v(\tau,y_i))$ \\
		$\mathcal{T}_{y_i} \leftarrow \mathcal{T}_{y_i} \cup \tau^k$ \\
	}
}

$\mathcal{T}^* = \{\mathcal{T}_{y_0}, \cdots, \mathcal{T}_{y_n}\}$

\end{algorithm}

\subsection{Stage 2: Policy Construction}
Policy construction is a process of mapping the evaluation results of stage 1 to the final augmentation policy for training target networks.
It is needed because augmentation policies are usually searched on light-weight proxy tasks such as density matching, but are evaluated on the complete tasks of image classification.
Even for methods that search on complete tasks such as AutoAugment \cite{cubuk2019autoaugment}, they still naively concatenate multiple searched policies into a final policy.
However, the policies for concatenation usually share a great potion of overlapped transformations, resulting in a high degree of redundancy.

In this paper, we propose an effective policy construction method to iteratively select candidate augmentation triples for the final policy, based on the mutual information criteria of minimum-redundancy maximum-relevance (mRMR) \cite{peng2005feature}.
Specifically, in \textit{LA3}, the relevance metric is defined as the predicted reward $\tilde{r}$ as it provides a direct evaluation on the performance of a certain augmentation triple.
And the redundancy of an augmentation triple $\tau$ is defined as the average number of intersecting operations between it and the already selected augmentation triples $\mathcal{T}_s$.
Formally, in each iteration of policy construction, we define the score $v(\tau, y)$ of each unselected augmentation triple $\tau$ at label $y$ as 
\begin{equation}
	\label{eq:score_function}
	v(\tau,y) = \tilde{r}_{\tau,y} - \alpha \times \overline{r} \times \frac{1}{|\mathcal{T}_s|} \sum_{\tau_s \in \mathcal{T}_s} |\tau \cap \tau_s|,
\end{equation}
where $|\tau \cap \tau_s|$ refers to the number of overlapped operations between $\tau$ and $\tau_s$, $\overline{r}$ is the average predicted reward of all augmentation triples in search space and is used to scale the redundancy, and $\alpha$ is a hyper-parameter adjusting the weight between the reward value and the redundancy value. 

\Cref{alg:stage2} illustrates the overall process of the policy construction stage where the goal is to find a label-aware policy containing a collection of augmentation triples that maximizes the rewards while keeping a low degree of redundancy.
Specifically, for each label $y_i$, we retrieve the predicted reward $\tilde{r}_{\tau, y_i}$ for each augmentation triple $\tau$ in the search space $A$.
Afterwards, a label-specific policy $\mathcal{T}_{y_i}$ is constructed iteratively by calculating the score $v(\tau, y_i)$ of unselected augmentation triples with \Cref{eq:score_function} and add the augmentation triple with the highest score to the policy until the required number of candidates $N_\text{cand}$ is met.
Eventually, the label-aware policy $\mathcal{T}^*$ is built with each label $y_i$ corresponding to a label-specific policy $\mathcal{T}_{y_i}$.

\section{Experiments}

In this section, we first describe the details of our experiment settings.
Then we evaluate the proposed method, and compare it with previous methods in terms of both performance and search cost.
Finally, we perform thorough analysis on the design of different modules in our algorithm.
Code and searched policies are released at \url{https://github.com/Simpleple/LA3-Label-Aware-AutoAugment}.

\subsection{Datasets, Metrics and Baselines}
Following previous work, we evaluate our \textit{LA3} method on CIFAR-10/100 \cite{krizhevsky2009learning} and ImageNet \cite{deng2009imagenet}, across different networks including ResNet \cite{he2016deep}, WideResnet \cite{zagoruyko2016wide}, Shake-Shake \cite{gastaldi2017shake} and PyramidNet \cite{han2017deep}.
Test accuracy is reported to assess the effectiveness of the discovered policies, while the cost is assessed by the number of GPU hours measured on Nvidia V100 GPUs.
For a fair comparison, we list results of stationary policies produced by static strategies, AutoAugment \cite{cubuk2019autoaugment}, FastAA \cite{lim2019fast}, and DADA \cite{li2020dada}.
We also include results from dynamic strategies, PBA \cite{ho2019population}, AdvAA \cite{zhang2019adversarial}, and MetaAug \cite{zhou2021metaaugment}, producing non-stationary policies as target model training progresses.

\subsection{Implementation Details}
\textbf{Policy Composition.}
For a fair comparison, we use the same $15$ augmentation operations as PBA and DADA do, which is also the same set used by AA and FastAA with SamplePairing \cite{inoue2018data} excluded.
Additionally, ``Identity'' operation that returns the original image is introduced in our search space to prevent images from being excessively transformed.
% We include in total $16$ image operations in our search space, including Identity, ShearX/Y, TranslateX/Y, Rotate, AutoContrast, Invert, Equalize, Solarize, Posterize, Contrast, Color, Brightness, Sharpness, and Cutout \cite{devries2017improved}.
Each label-specific policy consists of $N_\textit{cand}=100$ augmentation triples, while in evaluation, each sample is augmented by an augmentation triple randomly selected from the policy with random magnitudes. 

\textbf{Neural Predictor.}
The network structure of the neural predictor is composed of two embedding layers of size $100$ that map labels and augmentation operations to latent vectors and three fully-connected layers of hidden size $100$ with Relu activation function.
The representation of an augmentation triple is constructed by combining the three augmentation operation embedding vectors with mean-pooling and concatenating it with the label embedding vector.
Then it is passed into the FC layers to derive the predicted reward.
The predictor network is trained for $100$ epochs with Adam optimizer \cite{kingma2014adam} and a learning rate of $0.01$.

\textbf{Search Details.}
For CIFAR-10/100, we split the original training set of $50,000$ samples into a training set $D^{tr}$ of size $46,000$ to pre-train the model $\theta$, and a valid set $D^{val}$ of $4,000$ for density matching.
We search our policy on WRN-40-2 network and apply the found policy to other networks for evaluation.
For ImageNet, we randomly sample $50$ examples per class from the original training set, and collect $50,000$ examples in total to form the valid set, where the remaining examples are used as the training set.
In the augmentation exploration stage, the total number of iterations is set to $T=500$, and the warm-up iterations is set to $T_0=100$.
In the policy construction stage, $\alpha=2.5$ is used to calculate the reward values of augmentation triples.
% The computational cost reported in this paper is measured in GPU hours with NVIDIA V100 GPUs.

\textbf{Evaluation.}
The evaluation is performed by training target networks with the searched policies, and the results are reported as the mean test accuracy and standard deviation over three runs with different random seeds.
We do not specifically tune the training hyperparameters and use settings consistent with prior work.
We include the details in the supplementary materials.

\begin{table*}[tb]
	\centering

    \caption{Top-1 test accuracy (\%) on CIFAR-10 and CIFAR-100. We mainly compare our method \textit{LA3} with methods that also produce stationary augmentation policies, including AA, FastAA and DADA. Results of dynamic policies (PBA, AdvAA and MetaAug) are also provided for reference.}
    \label{tab:cifar-results}
    \resizebox{\textwidth}{!}{
      \begin{tabular}{ll|ccccc|ccc}
        \toprule
        \multirow{2}{*}{\textbf{Dataset}} & \multirow{2}{*}{\textbf{Model}} & \textbf{Baseline} & \textbf{AA} & \textbf{FastAA} & \textbf{DADA} &  \textbf{LA3} & \textbf{PBA} & \textbf{AdvAA} & \textbf{MetaAug}\\
        & & & static & static & static & static & dynamic & dynamic & dynamic\\
         
        \midrule
        CIFAR-10 & WRN-40-2 & $94.7$ & $96.3$ & $96.4$ & $96.4$ & $\mathbf{97.08\pm0.08}$ & $-$ & $-$ & $96.79$ \\
                 & WRN-28-10 & $96.1$ & $97.4$ & $97.3$ & $97.3$ & $\mathbf{97.80\pm0.15}$ & $97.42$ & $98.10$ & $97.76$ \\
                 & Shake-Shake (26 2x96d) & $97.1$ & $98.0$ & $98.0$ & $98.0$ & $\mathbf{98.07\pm0.11}$ & $97.97$ & $98.15$ & $98.29$ \\
                 & Shake-Shake (26 2x112d) & $97.2$ & $98.1$ & $98.1$ & $98.0$ & $\mathbf{98.12\pm0.08}$ & $97.97$ & $98.22$ & $98.28$ \\
                 & PyramidNet+ShakeDrop & $97.3$ & $98.5$ & $98.3$ & $98.3$ & $\mathbf{98.55\pm0.02}$ & $98.54$ & $98.64$ & $98.57$ \\
        \midrule
        CIFAR-100 & WRN-40-2 & $74.0$ & $79.3$ & $79.4$ & $79.1$ & $\mathbf{81.09\pm0.28}$ & $-$ & $-$ & $80.60$ \\
                  & WRN-28-10 & $81.2$ & $82.9$ & $82.8$ & $82.5$ & $\mathbf{84.54\pm0.03}$ & $83.27$ & $84.51$ & $83.79$ \\
                  & Shake-Shake (26 2x96d) & $82.9$ & $\mathbf{85.7}$ & $85.4$ & $84.7$ & $85.17\pm0.13$ & $84.69$ & $85.90$ & $85.97$ \\
                  & PyramidNet+ShakeDrop & $86.0$ & $\mathbf{89.3}$ & $88.3$ & $88.8$ & $89.02\pm0.03$  & $89.06$ & $89.58$ & $89.46$\\
        \bottomrule
      \end{tabular}
    }
\end{table*}

\subsection{Experimental Results}
\textbf{CIFAR-10/100.}
\Cref{tab:cifar-results} summarizes the CIFAR-10 and CIFAR-100 results of different auto-augmentation methods on a wide range of networks.
Among all static methods that produce stationary policies, \textit{LA3} achieves the best performance for all 5 target networks on CIFAR-10 and for 2 out of 4 target networks on CIFAR-100.
% , which evidently demonstrates the superiority of our proposed method.
When extending the comparison to also include dynamic strategies,  \textit{LA3} still achieves the best CIFAR-10 and CIFAR-100 accuracies on WRN-40-2, which is the original network on which policy search was performed. 
When transferring these augmentation policies found on WRN-40-2 to other target network models for evaluation, \textit{LA3} also achieves excellent performance comparable to the current best methods. In particular, \textit{LA3} achieves the highest score for WRN-28-10 on CIFAR-100.
These results evidently proves the effectiveness of \textit{LA3} as an augmentation strategy to improve model performance, and demonstrates the strong transferability of our label-aware policies across different neural networks.

\textbf{ImageNet Performance.}
In \Cref{tab:imgnet-results}, we list the top-1 accuracy of different methods evaluated on ResNet-50, as well as their computational cost. For a fair comparison, we also indicate whether the Batch Augment (BA) trick \cite{zhang2019adversarial}, which forms a large batch with multiple copies of transformed samples, is used for each method, with ``(BA)'' after the method name. 
We also indicate the number of transformations used in the batch augment.
Note that the search cost for dynamic methods is included in the training cost, since they learn a dynamic augmentation policy during the training of the target model. 
We include the results for \textit{LA3} both with and without batch augment.

From \Cref{tab:imgnet-results} we can observe that among all methods without the batch augment trick, \textit{LA3} achieves the best ImageNet top-1 accuracy of $78.71\%$, while the search only took $29.3$ GPU hours, which is $15$ times faster than FastAA. Although DADA is faster, \textit{LA3} is substantially better in terms of the ImageNet accuracy achieved.

Meanwhile, \textit{LA3 (BA)} achieves a new state-of-the-art ImageNet accuracy of $79.97\%$ surpassing all existing auto-augmentation strategies including dynamic strategies AdvAA and MetaAug, with a total computational cost $2$ times and $3$ times lower than theirs, respectively. 
The high cost of these dynamic policies is due to the fact that augmentation policies may vary for each sample or batch and must be learnt together with model training. 
By generating static policies, \textit{LA3} is a simpler solution that decouples policy search from model training and evaluation, which is easier to deploy in a production environment, without introducing specialized structures, e.g., the policy networks in AdvAA and MetaAug, into target model training.

\begin{table*}[tb]
\centering
  \caption{ResNet-50 top-1 test accuracy (\%) and computational cost on ImageNet. Batch Augment (BA) trick is used in the training of \textit{LA3} (BA), AdvAA (BA) and MetaAug (BA). The number of transformations used in batch augment is also given in the table.}
  \label{tab:imgnet-results}
  \resizebox{\textwidth}{!}{
  \begin{tabular}{l|ccccc|ccc}
    \toprule
    \multirow{2}{*}{\textbf{ }} & \textbf{Baseline} & \textbf{AA} & \textbf{FastAA} & \textbf{DADA} & \textbf{LA3} & \textbf{LA3 (BA)} &\textbf{AdvAA (BA)} & \textbf{MetaAug (BA)}\\
    & & static & static & static & static & static & dynamic & dynamic \\
     
    \midrule
    Batch Augment (BA) & n/a & n/a & n/a & n/a & n/a & $\times4$ & $\times8$ & $\times4$\\
    \midrule
    \textbf{ResNet-50 Acc (\%)} & $76.3$ & $77.6$ & $77.6$ & $77.5$ & $\mathbf{78.71\pm0.07}$ & $\mathbf{79.97\pm0.07}$ & $79.40$ & $79.74$\\
            %  & ResNet-200 & $78.5$ & $80.6$ & $80.6$ & $-$ & $81.32$ & $81.43$ & $-$ \\
    \midrule
    \midrule
    \textbf{Search Cost (h)} & $-$ & $15,000$ & $450$ & $1.3$ & $29.3$ & $29.3$ & $-$ & $-$ \\
    \textbf{Train Cost (h)} & $160$ & $160$ & $160$ & $160$ & $160$ & $640$ & $1,280$ & $1,920$ \\
    \midrule
    \textbf{Total Cost (h)} & $160$ & $15,160$ & $610$ & $161.3$ & $189.3$ & $669.3$ & $1,280$ & $1,920$ \\
    
    \bottomrule
  \end{tabular}
  }
\end{table*}

\subsection{Ablation Study and Analysis}
The reason of the success can be attributed to the following designs in our \textit{LA3} algorithm.

\textbf{Label-Awareness.}
One of the main contributions of the paper is to leverage the label information and separately learn policies for samples of different classes, which captures distinct characteristics of data and produces more effective label-aware policies.
The results of \textit{LA3} variant without label-awareness (i.e., searching for label-invariant policies) are shown in the first row of \Cref{tab:ablation-results}, which are constantly lower than \textit{LA3} in all experimental settings.
% This confirms that effectiveness of introducing the label-aware setting into augmentation search.
This confirms that label-aware augmentation policies are effective at improving target network accuracy. 

\begin{figure*}[t]
\centering
\begin{subfigure}{.32\linewidth}
  \includegraphics[width=\linewidth]{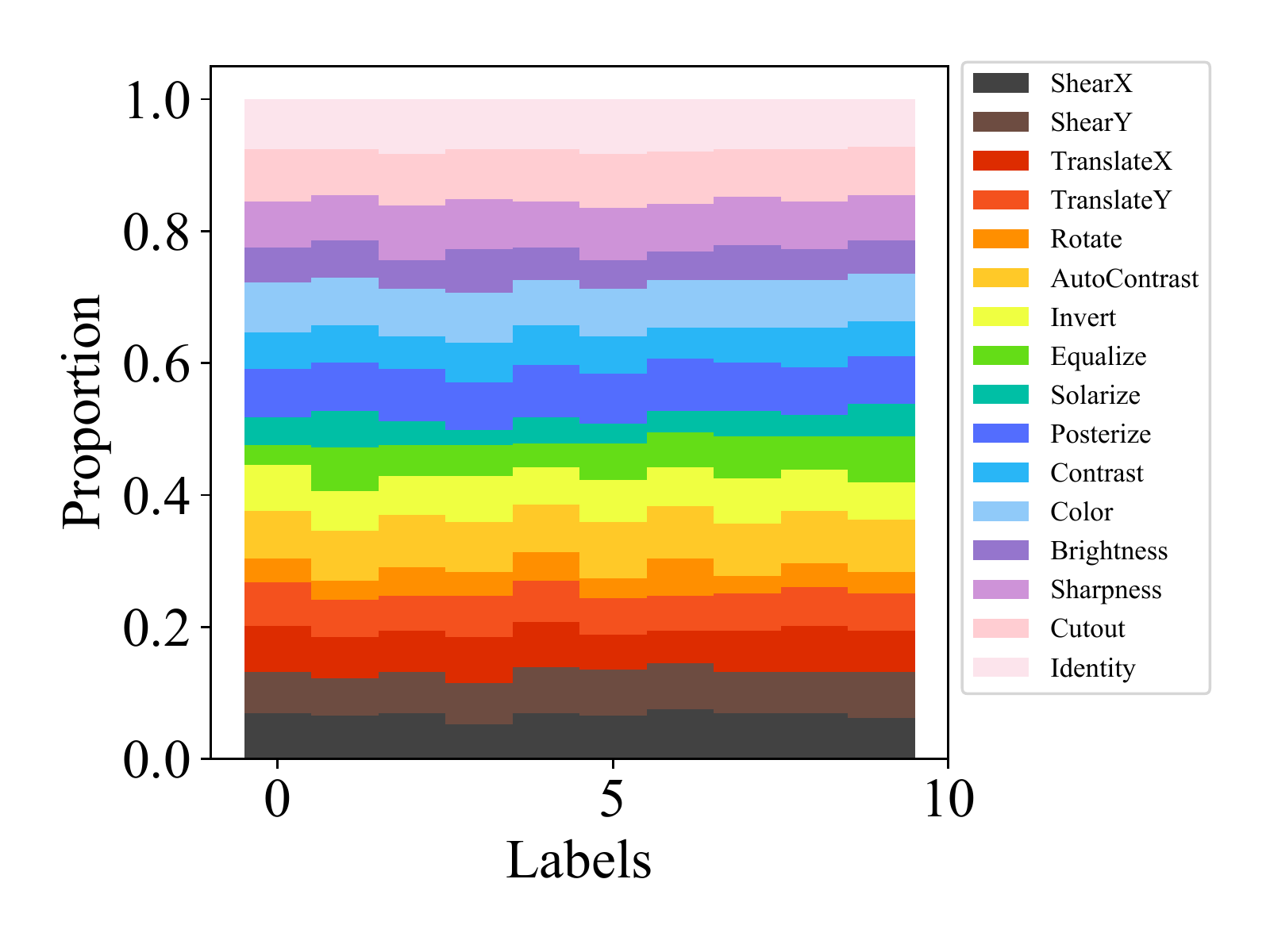}
  \caption{CIFAR-10 policy}
  \label{fig:cifar-10}
\end{subfigure}
\begin{subfigure}{.32\linewidth}
  \includegraphics[width=\linewidth]{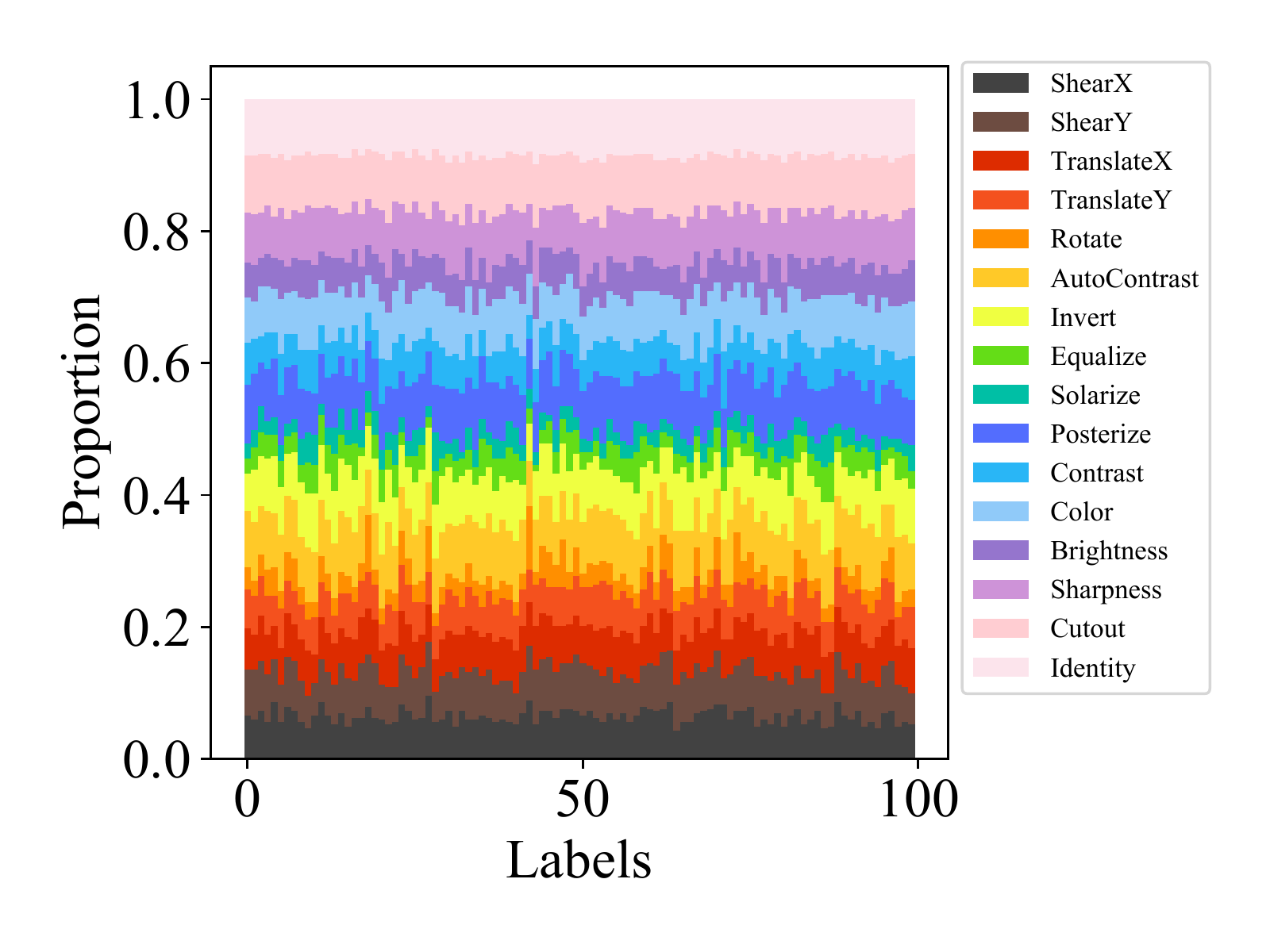}
  \caption{CIFAR-100 policy}
  \label{fig:cifar-100}
\end{subfigure}
\begin{subfigure}{.32\linewidth}
  \includegraphics[width=\linewidth]{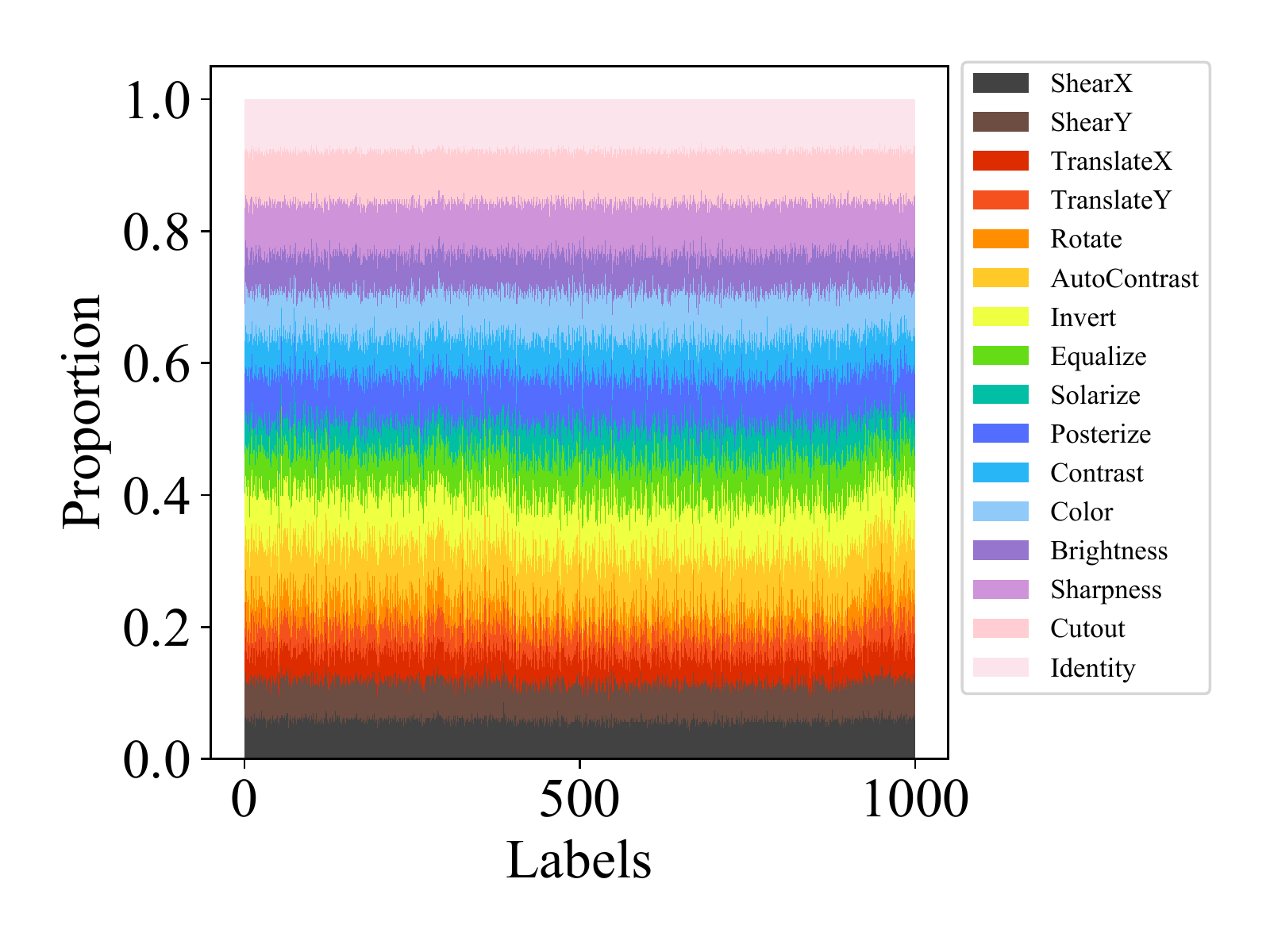}
  \caption{ImageNet policy}
  \label{fig:imgnet}
\end{subfigure}

\caption{The proportion of different augmentation operations in policies for different labels in \textit{LA3} searched label-aware policies on CIFAR-10, CIFAR-100 and ImageNet.}
\label{fig:Searched_Policies}
\end{figure*}

\begin{table}[tb]
  \centering
  \caption{Ablation analysis results in top-1 test accuracy (\%) on CIFAR-10 and CIFAR-100 with different designs removed from the full \textit{LA3} method.}
  \label{tab:ablation-results}
%   \resizebox{\columnwidth}{!}
%   {
  \begin{tabular}{l|cc|cc}
    \toprule
    \multirow{2}{*}{ }  & \multicolumn{2}{c|}{CIFAR-10} & \multicolumn{2}{c}{CIFAR-100} \\
    \midrule
    & WRN-40-2 & WRN-28-10 & WRN-40-2 & WRN-28-10 \\
    \midrule
    w/o Label-aware & $96.70$ & $97.11$ & $80.08$ & $82.76$ \\
    w/o Stage 2 (top-100) & $96.53$ & $97.49$ & $78.57$ & $82.76$ \\
    w/o Stage 2 (top-500) & $96.70$ & $97.26$ & $79.85$ & $84.04$ \\
    % w/o Top-100 & $96.53$ & $97.49$ & $78.57$ & $82.76$ \\
    % w/o Top-500 & $96.70$ & $97.26$ & $79.85$ & $84.04$ \\
    \midrule
    \textbf{LA3} & $\mathbf{97.08}$ & $\mathbf{97.80}$ & $\mathbf{81.09}$ & $\mathbf{84.54}$ \\
    \bottomrule
  \end{tabular}
%   }
\end{table}

\begin{figure}[t]
\centering
\begin{subfigure}{.4\linewidth}
  \includegraphics[width=\linewidth]{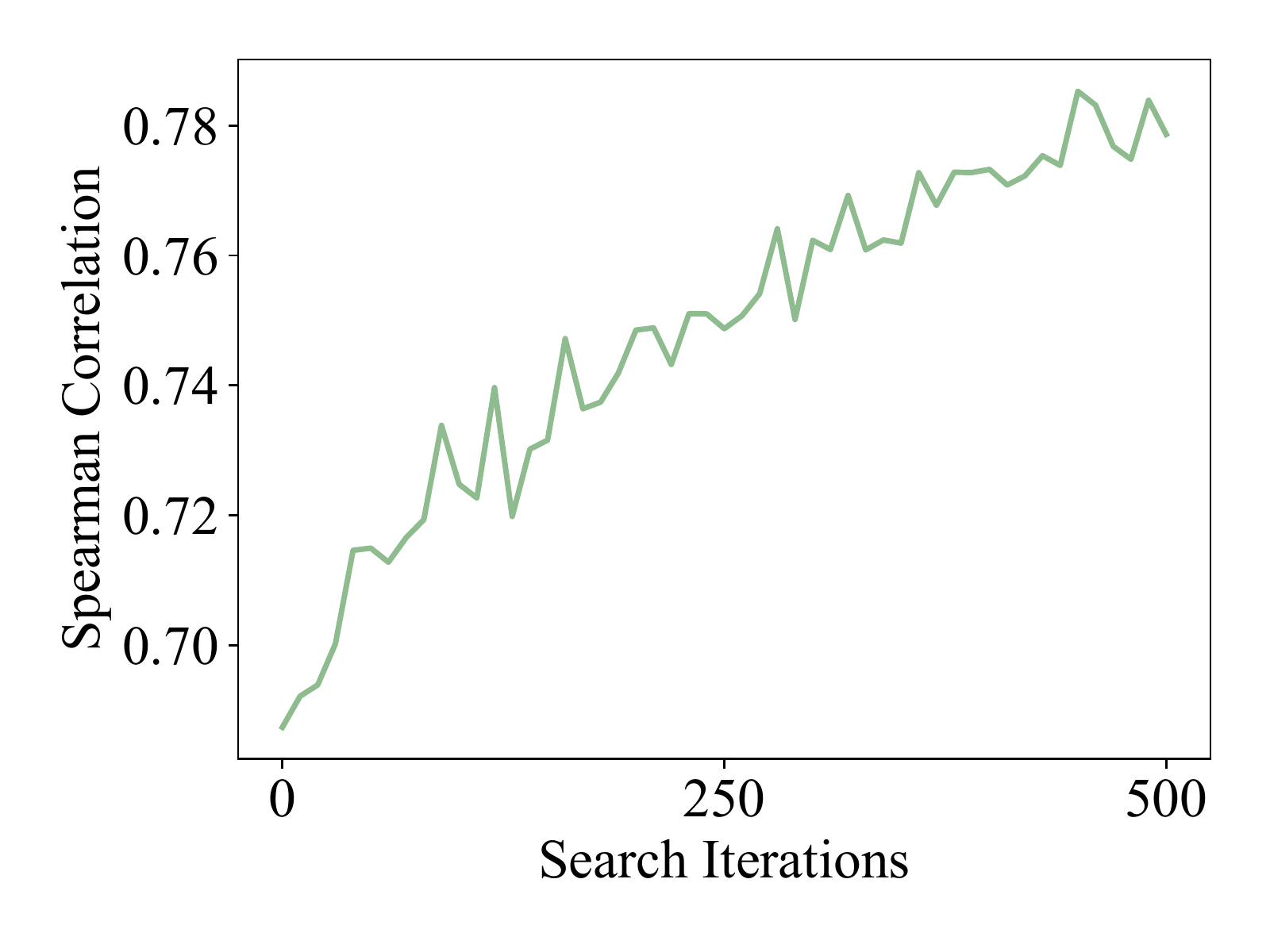}
  \caption{Spearman Correlation}
  \label{fig:predictor_spearman}
\end{subfigure}
\begin{subfigure}{.4\linewidth}
  \includegraphics[width=\linewidth]{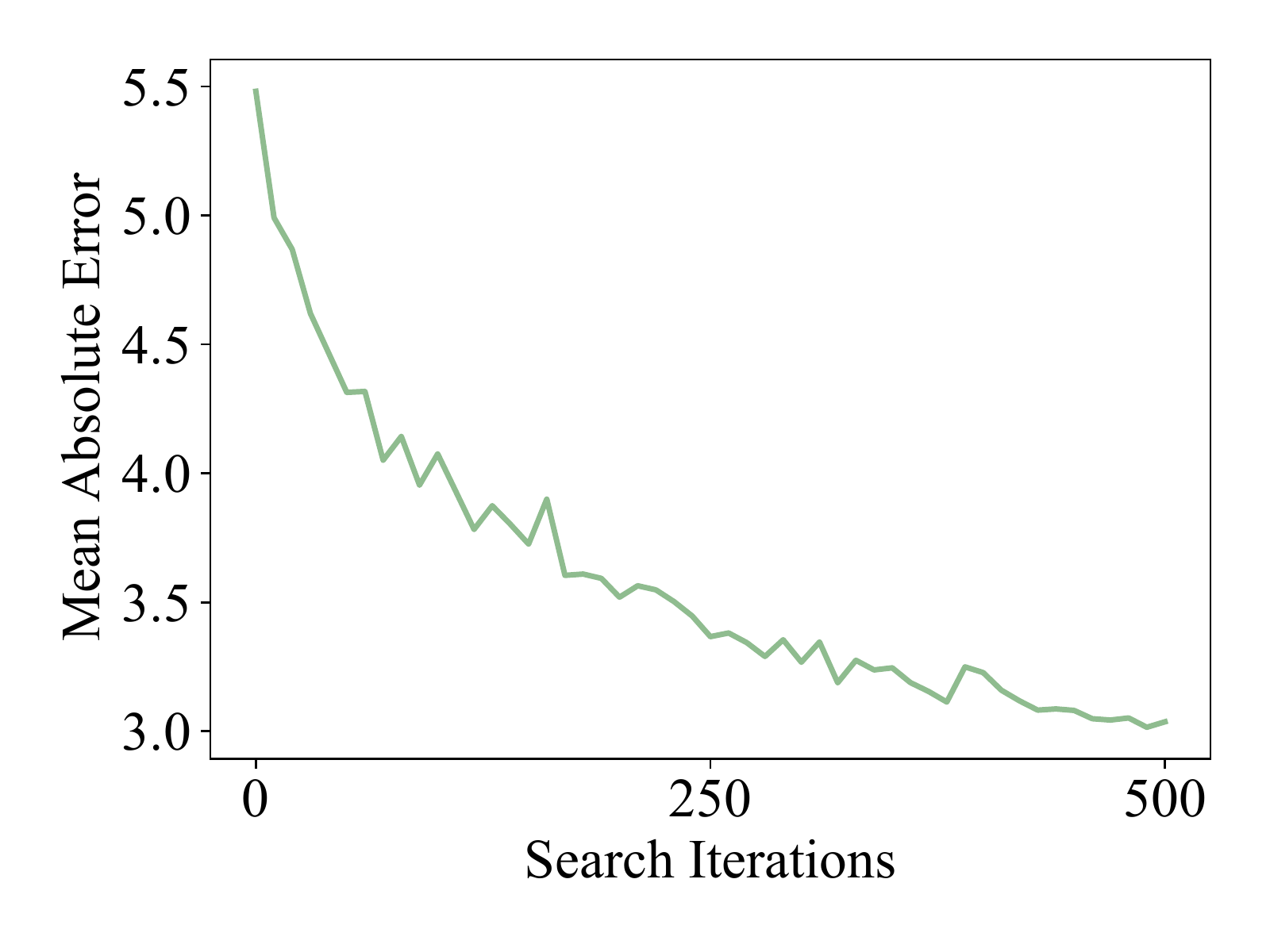}
  \caption{Mean Absolute Error}
  \label{fig:predictor_mae}
\end{subfigure}

\caption{The evaluation of the predictor during the policy search on ImageNet given by the Spearman's Rank Correlation and Mean Absolute Error over search iterations.}
\label{fig:Predictor_Res}
\end{figure}

\Cref{fig:Searched_Policies} gives an overview of the searched label-aware policies on CIFAR-10, CIFAR-100 and ImageNet, where we calculate the occurrences of different operations in each label-specific policy and plot their proportions in different colors. 
We can see that the derived policies possess a high diversity by having all the operations contributing to the final policy, meanwhile making the individual policies notably different among labels.
This observation further proves the need for separately treating samples of different labels in augmentation policy search.

\textbf{Neural Predictor.}
In addition to using density matching to simplify augmentation assessment during search, we have adopted a label-aware neural predictor to learn the mapping from an augmentation triple to its label-specific reward.
We now conduct a thorough evaluation to assess the performance of the neural predictor.
For each search iteration, the predictor is trained on $80\%$ of the history data and tested on the remaining $20\%$ data in terms of both the Spearman's Rank Correlation and Mean Abusolute Error (MAE). 
As shown in \Cref{fig:Predictor_Res}, as the policy search on ImageNet progresses and more samples are explored, the predictor can produce more accurate predictions of rewards, obtaining a $0.78$ Spearman Correlation and a decreased MAE when the search ends. This allows the predictor to properly guide the search process and find effective policies.

Furthermore, the use of the predictor better utilizes the search history and improves the sample efficiency during searching.
As a result, the search cost of our method is significantly reduced and is $15$ times lower than FastAA.

\textbf{Policy Construction.}
We evaluate the impact of our two-stage design on CIFAR-10 and CIFAR-100 datasets, by showing the performance of model variants with different policy construction methods in row 2 and 3 of \Cref{tab:ablation-results}.

We compare our policy construction method based on mRMR to the commonly used Top-k selection method adopted in AA \cite{cubuk2019autoaugment}, FastAA \cite{lim2019fast} and DADA \cite{li2020dada}.
We use two different $k$ value settings of $k=100$ equaling the number of candidates used in \textit{LA3}, and $k=500$ following the FastAA setting.
We can see that the policy that includes 500 augmentation triples per label with top predicted rewards yields a better performance than the policy with top 100 augmentation triples on both CIFAR-10 and CIFAR-100.
This can be attributed to the better diversity as more possibilities of augmentations are contained.
However, increasing the k value is not the best solution to improve augmentation diversity as the augmentation triples with high rewards tend to have similar compositions and may result in a high redundancy in the final policy.
Our \textit{LA3} incorporates a policy construction method that selects high-reward augmentation triples, and at the same time, keeping the lowest redundancy of the final policy.
With the two-stage design, our \textit{LA3} method beats the top-k variants and produces significant improvements in all settings.
% and produces significant improvements of $1.24\%$ and $0.50\%$ CIFAR-100 accuracy on WRN-40-2 and WRN-28-10, compared to top-500 results.

\textbf{Limitation.}
Unlike dataset-level augmentation policies that can be learned from one dataset and transferred to other datasets \cite{cubuk2019autoaugment,ho2019population,zhang2019adversarial}, \textit{LA3} learns label-aware policies where labels are specific to a dataset, and hence lacks the transferability across datasets, although \textit{LA3} demonstrates transferability across networks as shown in \Cref{tab:cifar-results}.
However, when dealing with a large dataset, \textit{LA3} can work on a reduced version of the dataset to search for label-dependent policies efficiently, and requires no tuning on training recipes when applying the found policy to the entire dataset.

\section{Conclusion}
In this paper, we propose a label-aware data augmentation search algorithm where label-specific policies are learned based on a two-stage algorithm, including an augmentation exploration stage based on Bayesian Optimization and neural predictors as well as a composite policy construction stage. 
Compared with existing static and dynamic augmentation algorithms, \textit{LA3} is computationally efficient and produces stationary policies that can be easily deployed to improve deep learning performance.
\textit{LA3} achieves the state-of-the-art ImageNet accuracy of $79.97\%$ on ResNet-50 among all auto-augmentation methods, at a substantially lower search cost than AdvAA and MetaAugment. 

\clearpage
\bibliographystyle{splncs04}
\bibliography{main}
\end{document}